\documentclass[final]{article}
\usepackage{nips_2017}
\usepackage[utf8]{inputenc} % allow utf-8 input
\usepackage[T1]{fontenc}    % use 8-bit T1 fonts
\usepackage{hyperref}       % hyperlinks
\usepackage{url}            % simple URL typesetting
\usepackage{booktabs}       % professional-quality tables
\usepackage{amsfonts}       % blackboard math symbols
\usepackage{graphicx} 
\usepackage{xcolor}
\usepackage{fontawesome}
\usepackage{graphicx}
\usepackage{fancyhdr}

\title{Tabular Embedding Model (TEM): Finetuning Embedding Models For Tabular RAG Applications}

\author{
  Sujit~Khanna \\ \texttt{sujit@synergia-ai.net} \\  \And Shishir~Subedi \\\texttt{shishirsubedi41@gmail.com} 
}

\date{\vspace{-5ex}} % Removing extra space below the author block

\begin{document}

\vspace{-20pt} 

\maketitle
\begin{center}
    \Large \textbf{Synergia}
\end{center}

\vspace{30pt} % Adds vertical space after the abstract

% \begin{abstract}
% \textbf{Abstract here \newline
% read similar white papers, aave, synthetix, curve, other dhedge}  
% \end{abstract}

\begin{abstract}
\label{sec:abstract}
In recent times Large Language Models have exhibited tremendous capabilities, especially in the areas of mathematics, code generation and general-purpose reasoning. However for specialized domains especially in applications that require parsing and analyzing large chunks of numeric or tabular data even state-of-the-art (SOTA) models struggle. In this paper, we introduce a new approach to solving domain-specific tabular data analysis tasks by presenting a unique RAG workflow that mitigates the scalability issues of existing tabular LLM solutions. Specifically, we present Tabular Embedding Model (TEM), a novel approach to fine-tune embedding models for tabular Retrieval-Augmentation Generation (RAG) applications. Embedding models form a crucial component in the RAG workflow and even current SOTA embedding models struggle as they are predominantly trained on textual datasets and thus underperform in scenarios involving complex tabular data. We choose the domain of financial markets for model evaluation to demonstrate TEM’s ability to handle intricate and high-dimensional datasets, an area where existing models typically falter. The evaluation results showcase that our approach not only outperforms current SOTA embedding models in this domain but also does so with a notably smaller and more efficient model structure.

\end{abstract}

{\section{Introduction}}\label{sec:intro}

Large Language Models have exhibited incredible capabilities in recent times across diverse sets of applications like mathematical reasoning  \cite{azerbayev2024llemma}, general-purpose problem-solving  \cite{openai2024gpt4}, \cite{guo2023evaluating}, \cite{murr2023testing} and code generation \cite{nijkamp2023codegen}. LLMs are also highly capable information or knowledge retrieval machines, however, they have some limitations in domain-specific tasks or knowledge-intensive applications. They are also notorious for producing hallucinations \cite{huang2023survey}, as they have training cut-off dates and their entire knowledge is abstracted in their trained weights. Since re-training LLMs is time-consuming and very costly the alternative is to provide the most recent/relevant information to the LLM in context via the Retrieval-Augmentation Generation approach (RAG) \cite{zhao2024retrievalaugmented} \cite{lewis2021retrievalaugmented}.In a generic RAG process given an input query, the retriever looks up the most relevant information in a knowledge base by semantic similarity search and sends it as additional context to the LLM to answer the query. This pipeline makes an LLM overcome its issues around domain-specific tasks and helps in accurately performing the tasks outlined by the user. 

The crucial components in RAG pipelines are the base LLM like GPT4, an embedding model, and a vector database. The embedding model is responsible for converting the external knowledge base into compressed embedding and is indexed into a vector database. Embedding models are also useful for retrieving the most relevant information from this vector store based on a user's query by doing similarity search. This makes the embedding model a crucial component of the RAG pipeline. The existing SOTA embedding models are trained dominantly on textual datasets which leads to sub-optimal performance on RAG applications that need tooling or numeric data. There are a few multi-task RAG approaches like LLM-Embedder, \cite{zhang2023retrieve} that are trained to retrieve examples, tools, and knowledge bases but we were unable to find any trained specifically for tabular data applications.

RAG applications in tabular data tasks are a niche but an important domain. Recent developments in tool usage have further amplified LLMs ability to analyze large-scale structured tabular data like CSV files or SQL tables. An example of such an application is asking LLM to \textbf{"Identify best performing stock by returns from S\&P500 index components over the last 6 months"}. For simpler tabular data tasks a generic embedding model suffices however when the nature of questions becomes complex; these models fail to retrieve the correct data chunks and begin hallucinating. 
% The task gets more daunting when the user's query needs tabular data across multiple tables/files and over a large range; which can stuff the 
In this paper we present our approach to solving this problem by finetuning a lightweight open-source embedding model for a tabular RAG task called Tabular Embedding Model or TEM. We present a unique embedding approach for tabular data that bypasses the need to embed the entire dataset, such an approach is highly scalable and can be applied to multiple tables and databases. We also outlined our RAG pipeline custom-built for this task by including a separate data analysis agent that only needs the relevant CSV files or tables from the retriever to execute the user's query. In this paper, we are focused on evaluating the model's performance in the domain of financial markets as it can encompass a diverse set of tasks from a host of interrelated and independent datasets. A semi-automated role-playing framework was used to generate the training and evaluation dataset from GPT-4. We found our finetuned model significantly outperforms existing SOTA embedding models for tabular data applications despite being much smaller in size.

{\section{Related Work}}\label{sec:Taxonomy}

\subsection{Table Retrieval and Q \& A}
Existing research in this space focuses on training LLMs directly on tabular data by handcrafting instruction finetuning datasets to imbibe it with tabular data understanding\cite{zhang2024tablellama}. Other approaches in this domain like \cite{liu2022tapex} \cite{yang2024unleashing} \cite{gong-etal-2020-tablegpt} perform pre-training on LMs like GPT-3 with mapping of natural language sentence and its corresponding table to the output of the answer. However, these approaches are not scalable when the analysis of tabular data runs into millions of rows. Another issue with a generalist tabular LLM is that the nuances involved in different applications are ignored especially in niche areas like financial markets. In recent times retriever-based approaches like \cite{chen2021open} proposed an approach based on early fusion and cross-block reader techniques for Q \& A tasks over both free text and tables. Other table retrieval research focuses on using intrinsic and extrinsic similarities for the retrieval process \cite{shraga-2020-ad}. This approach is further improved in T-RAG \cite{pan2022endtoend} that parses through a table corpus to directly locate the correct answer from the table cell. Such an approach creates obstacles for complex user queries when the solution to a problem requires analyzing millions of rows across a plethora of tables, this adds a lot of latency in the RAG process and creates issues in practical applications where speed of execution is crucial.

\subsection{Finetuning Embedding Models}
Recent advances in finetuning approaches have empowered LLMs to enhance their knowledge and understanding on less-popular or low-frequency concepts \cite{soudani2024fine} compares the RAG vs Finetuning approaches and identifies that both significantly enhance the capabilities of LLMs on domain-specific tasks. \cite{wang2024improving} exhibited how using a simple finetuning approach when used on an open-source decoder-only LLMs trained on a diverse set of synthetic datasets for hundreds of thousands of text embedding tasks across nearly 100 languages can achieve SOTA performance. Recently \cite{zhang2024mafin} introduced a novel approach of fine-tuning a black-box embedding model by augmenting it with a trainable open-sourced embedding model on domain-specific tasks. \cite{zhang2024raft} proposed an RAFT framework that combines RAG with Finetuning LLMs where given a question, and a set of retrieved documents, this approach trains the LLM to ignore those documents that don’t help in answering the question, this approach was shown to significantly improve the performance of 
LLM on domain-specific tasks. The above finetuning approaches are predominantly based on enhancing text-based retrieval and embeddings. A multi-application embedding model is proposed by \cite{zhang2023retrieve} that supports the diverse retrieval augmentation needs of LLMs with one unified embedding model that captures
distinct semantic relationships by modeling mutual interference across different tasks. However, we were unable to find any research that aimed at finetuning an embedding model specific to tabular data analysis applications. Most of the existing approaches when directly applied to such a task fail to generalize well and often run into scalability issues. The finetuning approach with the Tabular-RAG pipeline overcomes these problems and outperforms even the SOTA embedding models on our custom dataset relevant for financial tabular data analysis tasks.

% talk to data papers
%% https://arxiv.org/pdf/2302.11777.pdf
%% https://aclanthology.org/2023.nlp4convai-1.6.pdf
%% https://www.researchgate.net/publication/359647461_End-to-End_Table_Question_Answering_via_Retrieval-Augmented_Generation
%% https://arxiv.org/pdf/2203.16714.pdf
%% https://arxiv.org/pdf/2402.10666.pdf

%% https://arxiv.org/pdf/2005.08314.pdf

%% https://openreview.net/pdf?id=MmCRswl1UYl
%% https://arxiv.org/pdf/2306.11843.pdf

%% https://arxiv.org/pdf/2205.09843.pdf

% Finetuning papers
%% https://arxiv.org/abs/2311.09206
%% https://arxiv.org/pdf/2401.00368.pdf
%% https://arxiv.org/abs/2310.09263
%% https://arxiv.org/pdf/2403.01432.pdf
%% https://arxiv.org/abs/2403.10131
%% https://arxiv.org/pdf/2403.01432.pdf
%% https://arxiv.org/html/2401.00368v1

% Multi-model RAG:
%% https://arxiv.org/pdf/2310.07554.pdf

{\section{TEM (Tabular Embedding Model)}\label{sec:Eval}

In this section, we describe the TEM a new approach to finetuning smaller open-sourced embedding model that is trained on general language corpus for a sophisticated tabular RAG application. We first introduce the classical workflow of using embedding models in tabular rag applications, present our RAG workflow and indexing strategy for tabular data applications, describe a semi-automated dataset-generating process, and finally present a novel finetuning framework for tabular embedding model.

% like MSMARCO, NQ, QRecc as well as Non-labeled and instruction finetuning datasets like ArXiv, CodeParrot and Flan \textbf{Add references to the datasets} 
% \textbf{(explain what bge is trained on: https://arxiv.org/pdf/2310.07554.pdf Labeled data: (MSMARCO, NQ), QReCC and Non-labeled datasets and instruction finetuning datasets)}. 

% We finetuned this embedding model for sophisticated tabular RAG application using our custom dataset and training framework. In this section we first introduce the 

% \begin{itemize}
%     \item classic workflow of using embedding models in tabular rag applications
%     \item  present our optimized (change word) process of indexing the tabular data
%     \item RAG workflow for tabular data applications
%     \item Training data and its DGP
%     \item Finetuning approach for a open sourced embedding model    
% \end{itemize}

\subsection{Embedding tabular data in RAG pipeline}
A traditional embedding process for indexing the non-tabular data consists of creating chunks of documents, creating embeddings via a Embedding Model like OpenAI's text-embedding-3-large, and then indexing and storing them in a vector database. This approach fails for tabular data like CSV files or an SQL table because 

\begin{itemize}
    \item Tabular data can run into billions of rows, where creating chunks adds a lot of scalability issues
    \item Sending relevant chunks to the LLM at the end of the RAG process will quickly stuff the context window, which would make any analysis of the relevant data useless
    \item Creating chunks adds a lot of redundancies, where the majority of chunks are irrelevant to a user's question
\end{itemize}

% A work around to this approach is to link RAG pipeline to an another agent responsible for analyzing the data, this circumvents the need to add full dataset to the context window of the LLM. We limit the data required for indexing by only adding information related to particular csv file - which consists of csv filename, filename description, it's column names and definition of those column names. This process is achieved first by creating a mapping template for each csv filename. Where the filename is a key and its metadata is stored as the value in a dictionary. 

% \textbf{\textit{Integrate below sentence in above description}}

% For equities since content only differs by ticker symbol we made single chunks for each of daily, intraday and options factors with {SYMBOL} placeholder to be replaced. Then we loop through all the equities files and replaced that {SYMBOL} placeholder with actual ticker symbol name from csv filename itself there by making chunks for all equities files.

Figure 1 contrasts the tabular data indexing approach vs generic approach.

\begin{figure}[t]
\begin{center}
  \includegraphics[scale=0.35]{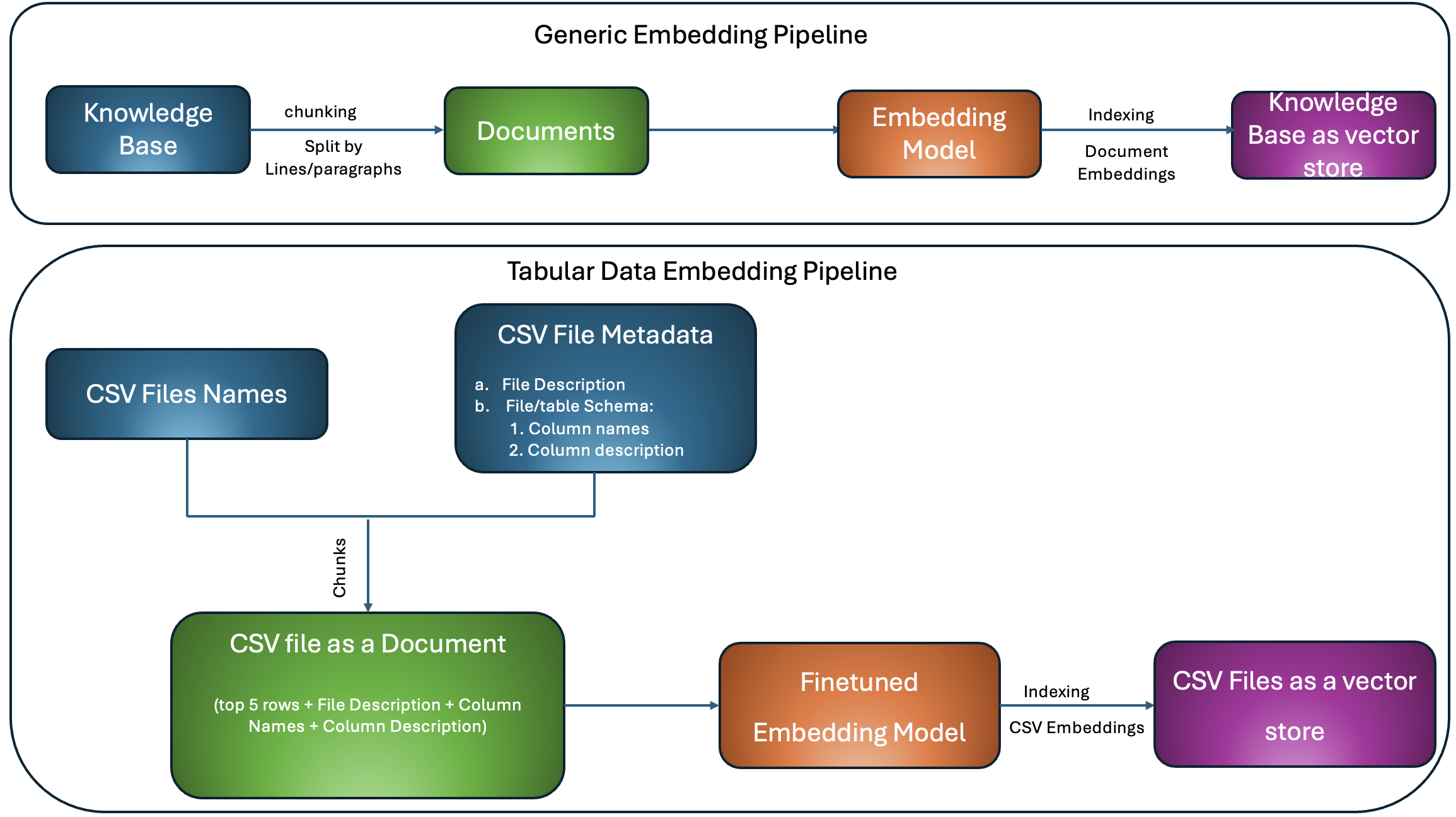}
  \caption{Generic indexing approach vs Tabular Embedding Approach}
  \label{fig:system}
  \end{center}
\end{figure}

% \begin{figure}[!htb]
% \minipage{0.3\textwidth}
% \includegraphics[scale=0.3]{generic_embedding.png}
% \caption{BCW SGD Loss}\label{fig:sgd_loss_wisc}
% \endminipage\hfill
% \minipage{0.7\textwidth}
% \includegraphics[scale=0.4]{Tabular_embedding.png}
% \caption{BCW MMO Loss}\label{fig:mmo_loss_wisc}
% \endminipage\hfill
% \end{figure}
% \clearpage

\subsection{RAG Pipeline}
As mentioned in the previous subsection for efficient performance of RAG on tabular data tasks, we need to integrate a data analysis agent into our workflow. Doing this reduces the burden on the RAG pipeline to ingest the entire dataset for analysis. Instead, the task of the RAG pipeline is to recommend the most relevant CSV files/tables based on the user questions. The data analysis agent built on top of the LLM then uses the most relevant file and performs the data analysis requested by the user. This pipeline is described in more detail in Figure 2 and crucial components of this pipeline are \newline

\begin{itemize}
    \item \textbf{Context:} This component provides additional context to the base LLM that enables it to pick the most appropriate set of files from the output of the embedding model and generate the code relevant to the user's query

    \item \textbf{Code Evaluator:} This component evaluates the code from the base LLMs and provides feedback to the LLM to modify the code if it has any errors or mistakes. This is done via a built-in reflection framework to identify any possible errors, and debug and resolve them to create accurate and optimized code.

    \item \textbf{Code Executor:} As the name suggests this component is responsible for executing the final version of the code

    \item \textbf{Final Response:} This component is responsible for deciding how the final output of the code should be saved or displayed for the user

\end{itemize}

% \textbf{Please add more description using the figure shown below, add as much detail as possible: Explain other sections in detai that are not a part of the above description}

\begin{figure}[t]
\begin{center}
  \includegraphics[scale=0.35]{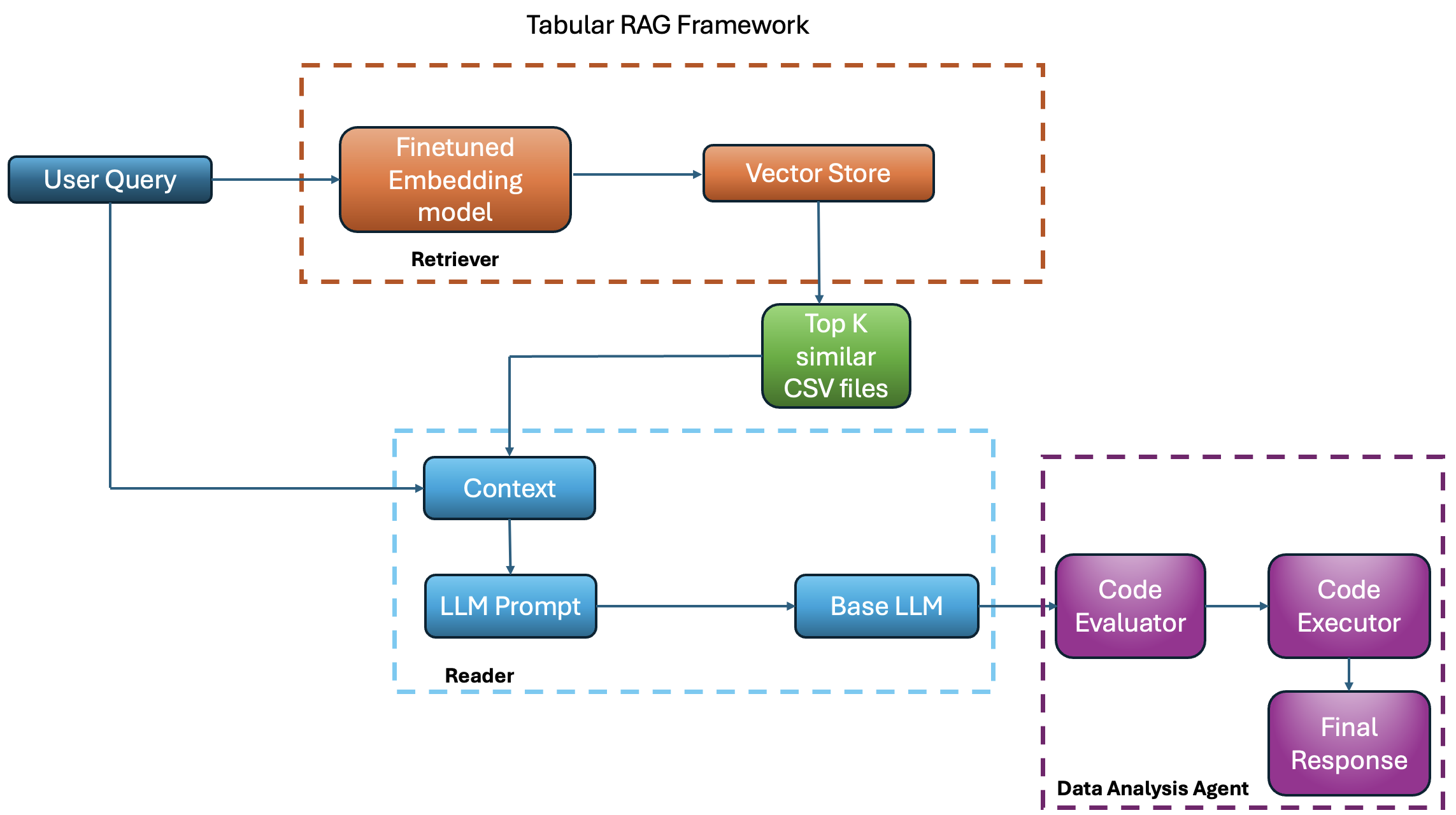}
  \caption{Proposed Tabular RAG Workflow}
  \label{fig:system}
  \end{center}
\end{figure}

\subsection{Training data for finetuning}
Our research is primarily focused on tabular data analysis in financial markets, we chose data from this domain as there are a plethora of data sources that can explain different market effects. There are plenty of financial instruments and asset classes across which we can analyze the data and financial datasets are highly dimensional and non-linear, which adds another layer of complexity to tabular data analysis tasks.\newline

% \textbf{\textit{Please restructure above sentence to make it more precise}}

Our dataset consists of questions across many stocks, macro variables, and data from options markets as well as from other asset classes like commodities, bonds, etc. Specifically, our dataset contains a mapping of
questions to relevant files, where each question can be mapped to $[1,..n]$ files where $0<n<=5$ files. Files can belong to the same subset i.e. all files belonging to equities or any other combination of asset classes. 

\begin{figure}[t]
\begin{center}
  \includegraphics[scale=0.35]{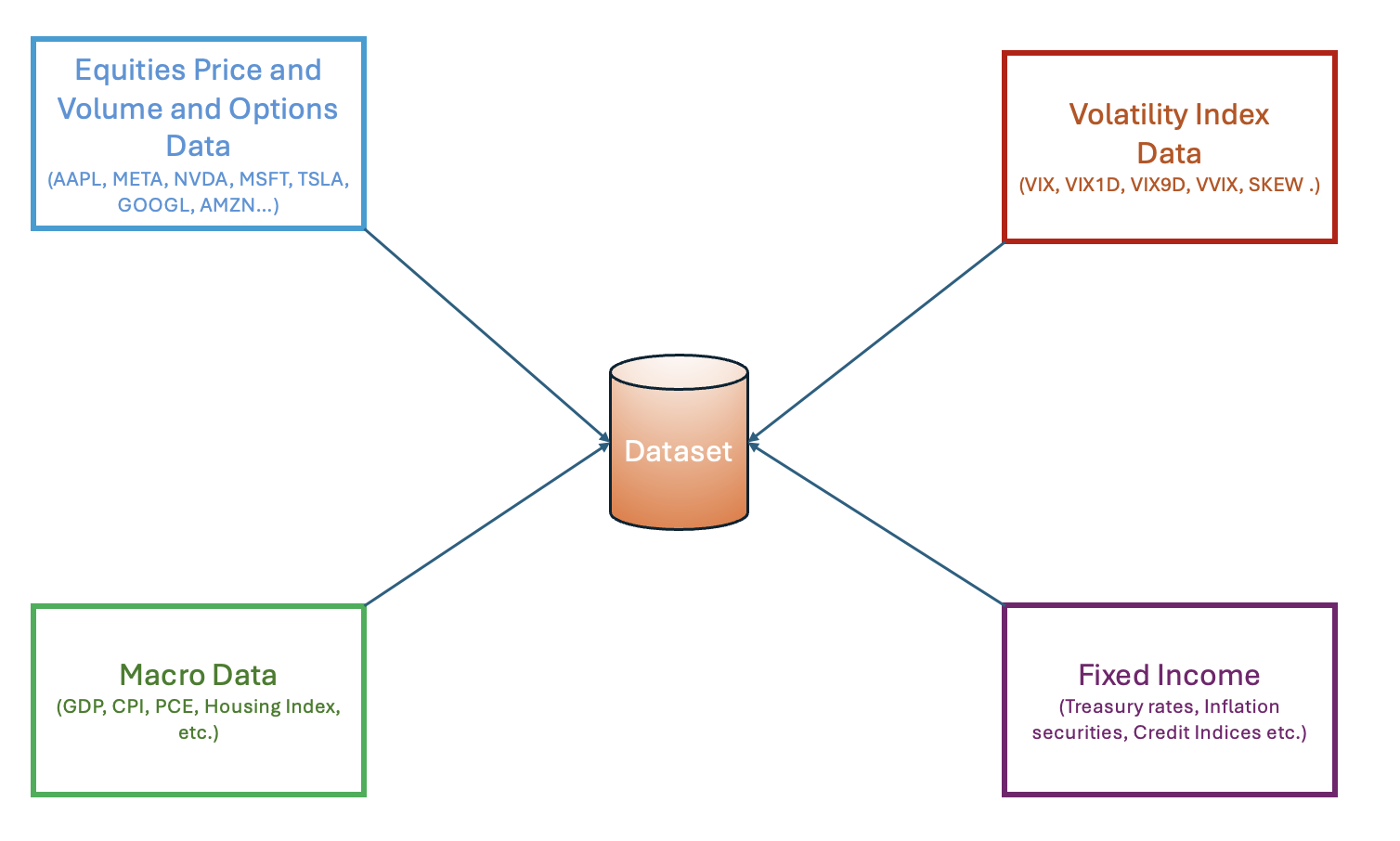}
  \caption{Custom finetuning dataset}
  \label{fig:system}
  \end{center}
\end{figure}

We use a semi-automated process of generating the dataset using a mixture of role-playing and a few shot prompts to generate a diverse set of questions. We initially provided the following information in context to the LLM. (We used GPT4 for generating the dataset)

\begin{itemize}
    \item A dictionary mapping file name to its description and column names with its definitions
    \item Guidelines for generating the questions i.e. specific set of rules to follow in the data generation process
    % \item A host of few shot prompt examples
    % \item A prompt template asking the LLM to think like a researcher
\end{itemize}

The next step involves a role-play-based approach to generate a diverse set of questions. In this role-play, the user asks LLM to assume multiple roles such as a fundamental analyst, macro trader, machine learning expert, data scientist as well as a retail trader. A prompt template is used for role-playing which also provides a few shot examples to generate a diverse set of questions. This process is described in Figure 3. In this role play, we also adjust for the number of relevant files required for each question. The hierarchical flow of such an approach is shown in Figure 4.

\begin{figure}[t]
\begin{center}
  \includegraphics[scale=0.5]{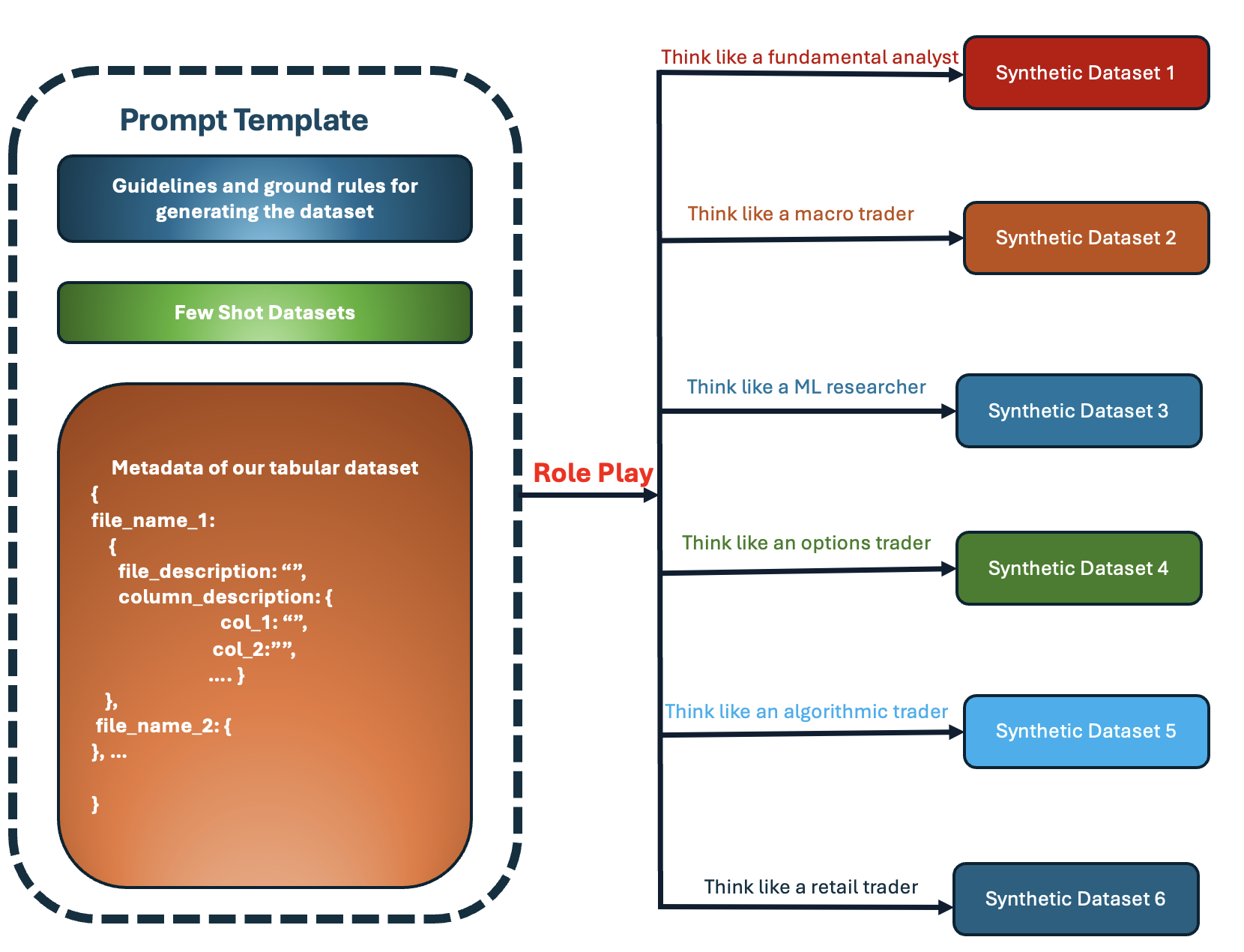}
  \caption{Semi-automated data generation process via role-play}
  \label{fig:system}
  \end{center}
\end{figure}

% \textbf{TODO: Generate a table containig the stats of the generated questions, breakdown by n, by asset-classes, etc }

\begin{figure}[t]
\begin{center}
  \includegraphics[scale=0.4]{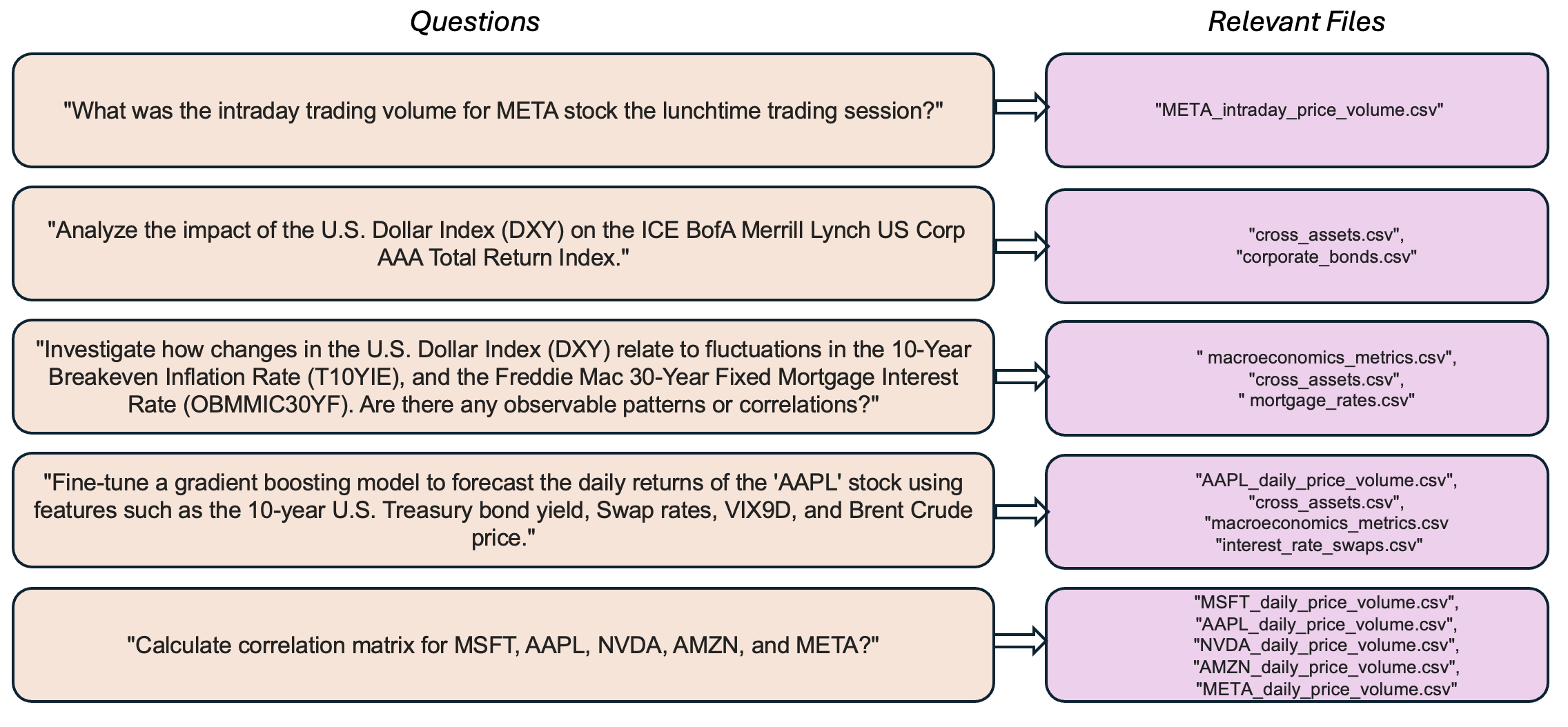}
  \caption{Sample Questions generated via role-playing framework}
  \label{fig:system}
  \end{center}
\end{figure}

\subsection{Finetuning methodology}
Our finetuning methodology consists of 2 steps, in the first step we initialize the base embedding model with "New Word Embeddings" \cite{hewitt2021initializing}, and the second step involves finetuning our model with the dataset generated in the first step

\subsubsection{New words embedding initialization}
The current finetuning approaches mostly use instruction finetuning datasets to train a language model for a downstream task. A better approach is to first expand the vocabulary of the model by adding new words.
This can be done by resizing the base model’s embeddings to make new embeddings for the words in the updated vocabulary i.e. vocabulary for the downstream task. \newline However this approach poses a problem during the training process where sampling from the existing distribution of reduces the probabilities of words post-expansion. More specifically 
Let $w_{i:n}$ be a sequence of words in the vocabulary $V={1....n}$. Let our language model be characterized as \newline

% $P_{\theta}(w_{i}/w_{1:i-1}) = \frac{e^{h_{i-1}^{t} m_{w_i}}}{\sum_{j=1}^{n} e^{h_{i-1}^{t}}^{m_{j}}}$
$P_{\theta}(w_{i}/w_{1:i-1}) = \frac{e^{h_{i-1}^t m_{w_i}}}{\sum_{j=1}^n e^{h_{i-1}^t m_j}}$ where 
% $h_{i-1}=\phi_{\theta}(w_{1:t-1}) \in R_^{d}$ 
$h_{i-1}=\phi_{\theta}(w_{1:t-1}) \in \mathbb{R}^d$
and $m_{i} \in R^{d}$ is the embedding for word i. Post expansion when we add new words to vocal $n + 1 \not\in V$, we sample new word embedding ${m_{n + 1}}$ from the initial distribution. \newline So our new LM is characterized as  

$P_{\theta^{i\textquotesingle }}(w_{i}/w_{1:i-1})$, where $\theta^{\textquotesingle} = \theta \cup \{m_{n + 1}\}$ such that,  \newline

% $P_{\theta}(w_{i}/w_{1:i-1}) = e^{h_{i-1}^{t}}m_{w_i}/\sum_{j=1}^{n} e^{h_{i-1}^{t}}m_{j}$ where \newline
% $h_{i-1}=\phi_{\theta}(w_{1:t-1}) \in R_^{d}$ \newline
% $m_{i} \in R^{d}$ is the embedding for word i

% $\frac{e^{h_{i-1}^{t}}m_{w_i}}{\sum_{j=1}^{n} e^{h_{i-1}^{t}}m_{j}}$

% Post expansion when we add new words to vocal $n + 1 \not\in V$, we sample new word embedding ${m_{n + 1}}$ from the initial distribution. So our new LM is characterized as  $P_{\theta^{i\textquotesingle }}(w_{i}/w_{1:i-1})$, where $\theta^{\textquotesingle} = \theta \cup \{m_{n + 1}\}$ such that \newline

% $P_{\theta^{\textquotesingle }}(w_{i}/w_{1:i-1}) = \frac{e^{h_{i-1}^{t}}^{m_{w_i}}}{( Z + \sum_{j=1}^{n} e^{h_{i-1}^{t}}^{m_{n + 1}})} $

$P_{\theta'}(w_{i}/w_{1:i-1}) = \frac{e^{h_{i-1}^t m_{w_i}}}{Z + \sum_{j=1}^{n} e^{h_{i-1}^t m_{n+1}}}$. \newline
The relation between pre and post-expansion LM can therefore be represented as,

$P_{\theta'}(w_{i}/w_{1:i-1}) = P_{\theta}(w_{i}/w_{1:i-1}) \times \frac{1}{1 + \frac{e^{h_{i-1}^t m_{n+1}}}{Z}}
$ \newline

% $Z = \sum_{i}^{n} e^{h_{i-1}^{t}}m_{w_j}$ is the partition function of pre-expansion LM. \newline

% $P_{\theta^{\textquotesingle }}(w_{i}/w_{1:i-1}) = e^{h_{i-1}^{t}}m_{w_i}/ ( Z + \sum_{j=1}^{n} e^{h_{i-1}^{t}}m_{n + 1})$  \newline
% $Z = \sum_{i}^{n} e^{h_{i-1}^{t}}m_{w_j}$ is the partition function of pre-expansion LM. \newline

% $P_{\theta^{\textquotesingle }}(w_{i}/w_{1:i-1}) = P_{\theta}(w_{i}/w_{1:i-1}) \times 1/((1 + e^{h_{i-1}^{t}}m_{n + 1})/Z)$ \newline

From the above relation if Z is small relative to the new distribution, then the probabilities of all pre-expansion words decrease a lot. The KL divergence between pre and post-word expansion is \newline

% $KL (p_{\theta}^{\textquotesingle} \vert \vert p_{\theta}) = log(  1 + \frac{(1 + e^{h_{i-1}^{t}}^{m_{n + 1}})}{Z})$

$KL(p_{\theta'} \parallel p_{\theta}) = \log\left(1 + \frac{1 + e^{h_{i-1}^t m_{n+1}}}{Z}\right)$

This can cause huge divergences between pre and post-expansion models, especially in case of small norm initialization. A simple way to circumvent this is to average all initial embeddings and instantiate new word embeddings with them. Formally we have \newline

$\mu = 1/n \times \sum_{i}^{n} m_{i}$ \newline
$ \sigma = 1/n \times (M - \mu)^{T}(M - \mu)$ \newline
$m_{n + 1} \sim \mathcal{N(\mu, \sigma)}$ \newline

This change would bound the KL divergence to $KL (p_{\theta}^{\textquotesingle} \vert \vert p_{\theta}) = log (\frac{1}{ 1+ \frac{1}{n}})$.As an added advantage the bounded probabilities are that gradients are well-behaved as the norm of the gradient balances extreme probabilities.

% datasets adds new word to the vocabulary of the embedding model, especially if we're finetuning a generic embedding model for a niche task, as described in this paper. In this approach when we add new untrained word embeddings to the model we lose the distribution of the pretrained model which increases the KL-divergence between and pretrained and finetuned model distribution. More specifically \newline ${add_johns_article_math here}$

% Instead if during the model initialization if we average the   word embeddings of the extended vocabulary in the embedding model we can bounds the KL-divergence between the pre-expansion and post-expansion language models’ token-level distributions. This process improves the finetuning process. ${add_more_of_johns_article_math here}$

\subsubsection{Finetuning the model}

We use BGE-lage-en-v1.5 as our base model on top of which we perform new words embedding initialization explained in the previous step. The dimensions and sequence length for this model are 1024 and 512 respectively. Our reasons for using this model were two-fold,

\begin{itemize}
    \item We wanted to identify a light embedding model that could be trained on a local laptop
    \item Model that is comparable in performance with other SOTA models.
\end{itemize}

% The table below shows the performance of BGE-large-en on major benchmark datasets from huggingface. 

% \textbf{TODO: add an image from huggingface leader board showing the position of bge-large model}

As explained in previous sections our dataset consists of mapping of questions to relevant files, such that $q \mapsto \{p \mid 1 \leq p \leq 5\}$. Figure 5 shows the examples for each value of n.

We use multiple negative ranking (MNR) loss functions in our framework as it's proven its performance in information retrieval tasks. This loss function is also helpful for training sets with only positive pairs as in our case, also known as anchor positive pairs. This loss function uses each example in a mini-batch as a negative sample for all other examples. For a given anchor-positive pair, the loss is calculated based on the similarity of the anchor to its positive pair and the negative samples (other examples in the batch). i.e. The loss function promotes learning embeddings where questions are closer to their corresponding relevant context and farther from other contexts in the batch

% \textbf{restructure above sentence} \newline

The following steps are involved in the finetuning process

\begin{itemize}
    \item \textbf{Batch Preparation: } For each training step, you prepare a batch of question-context pairs. Let's assume the batch size is $N$

    \item \textbf{Loss Calculation:} The loss for each question $q_{i}$ with its relevant mapping of files as context $p_{i}$ in the batch is computed against all other contexts in the same batch, treated as negative examples: \newline

    $L(q_i, p_i) = log (\frac{e^{sim(q_i,p_i)}}{\sum_{i=1}^{N} e^{sim,(q_i,p_i)}}) $, where $sim(q_i, p_i)$ denotes the cosine similarity score between the between the embeddings of question $q_{i}$ with its relevant mapping of files as context $p_{i}$

    \item \textbf{Average Loss:} The average loss over the batch is calculated, which we aim to minimize \newline

    $L_{batch} = \frac{1}{N}\sum_{i=1}^{N} L(q_i, p_i) $
    
\end{itemize}

Our optimization framework uses AdamW with a linear warmup scheduler. AdamW optimizer exhibits superior performance in finetuning applications \cite{kingma2017adam}, and is better equipped to deal with sparse gradients than other optimizers.

$W_{t + 1} = W_{t + 1}^{\textquotesingle} + \eta \lambda W_{t}$ \newline  $W_{t + 1}^{\textquotesingle} = W_{t} - \eta*B_{t}$ \newline Where $W_{t + 1}^{\textquotesingle}$ is Weight Update Without Weight Decay and $B_{t}$ is bias correction factor.

The linear warmup scheduler increases the learning rate $\eta$ linearly during the warmup phase and then decays $\eta$ linearly to zero 

Additionally, we use a batch size of 5 and finetune the model for 50 epochs. Because of the lightweight nature of this model, we were able to finish finetuning this model on a Macbook M3max chip with 64GB RAM and 40 core GPU in little less than 8 hours.

{\section{Evaluation}

Due to the niche nature of the task; we use the test set created using the same role-playing approach described in the previous section.

\subsection{Baseline models}
For benchmarking the performance of our approach we use the best-in-class embedding models from both open and proprietary sources described below.

\begin{itemize}
    \item \textbf{SFR-Embedding-Mistral}: An embedding model by salesforce which is built on top of E5-mistral-7b-instruct and Mistral-7B-v0.1 models. It has embedding dimensions of 4096, a context window of 32768, and a size of 14.22 GB
    
    \item \textbf{text-embedding-3-large}: This is the most recent and best-performing embedding model from openAI. It has embedding dimensions of 3072 and a context window of 8191. Its size is unknown as it's a proprietary OpenAI model
    
    \item \textbf{BGE-large-en-v1.5:} This is a general-purpose embedding model published by BAAI. It has embedding dimensions of 1024, a context window of 512, and has a size of 1.34 GB
\end{itemize}

\subsection{Results}
% The breakdown of our test set is shown below \textbf{describe how test set is composed}

We analyze the results of the 4 models across a host of metrics relevant to our task i.e. we only include metrics that can assess the 
performance for our specific retrieval task. We have excluded metrics like Mean Reciprocal Rank (MRR) and Normalized Discounted Cumulative Gain (nDCG), as the performance of the downstream task by the agent is dependent on getting the correct top-k set of files while ignoring its ranking. Looking at the table we can see our finetuned model significantly outperforms the benchmark models across all metrics. The performance is relatively better for n<=3, which can be attributed to smaller embeddings and the context window of our model. 

% \textbf{Identify other patterns in the results and explain/analyze their reasons}.  \newline

% \textbf{identify why certain benchmark models fail at more complex tasks}

Embedding models are generally trained on a large and diverse corpus of text data; the nuances and patterns of tabular and numeric data are generally ignored in the datasets. For domain-specific applications like tabular RAG, this generally leads to sub-optimal performance. The table below outlines the results of 3 baseline models against our finetuned model across precision@10, recall@10, and hit rate. We used a value of $k=10$ as our applications generally need files in the range $1<=n<=6$ this k value should be sufficient to include all relevant files. In this study, we're ignoring the rank orders of the relevant files as the data science agent only needs relevant files and does not care about rank ordering.

\begin{table}[!htbp]
\centering
\caption{Tabular RAG Performance}
\label{results:single_v_mmo}
\scalebox{0.8}{
\begin{tabular}{*5c}
\toprule
Evaluation Metric & SFR-Embedding-Mistral & text-embedding-3-large & bge-large-en-v1.5 & bge-large-en-v1.5-finetuned  \\

Precision@10 & 0.2025 & 0.2041 & 0.1702 & \textbf{0.2160} \\
Recall@10 & 0.7477 & 0.7578 & 0.6643 & \textbf{0.7989} \\
Hit Rate@10 & 0.3892 & 0.3984 & 0.3100 & \textbf{0.4420} \\
\midrule

\bottomrule
\end{tabular}}
\end{table}

% \begin{table}[!htbp]
% \centering
% \caption{Accuracy breakdown with N value}
% \label{results:single_v_mmo}
% \scalebox{0.8}{
% \begin{tabular}{*5c}
% \toprule
% N & SFR-Embedding-Mistral & text-embedding-3-large & bge-large-en-v1.5 & bge-large-en-v1.5-finetuned  \\

% 1 & 0.7727 & 0.7784 & 0.75577 & \textbf{0.7955} \\
% 2 & 0.7143 & 0.6786 & 0.5982 & \textbf{0.8482} \\
% 3 & 0.2823 & 0.3629 & 0.1774 & \textbf{0.3871} \\
% >3 & 0.1243 & 0.1301 & 0.0376 & \textbf{0.1503} \\

% \midrule
% \bottomrule
% \end{tabular}}
% \end{table}

\begin{table}[!htbp]
\centering
\caption{Hit Rate@10 breakdown with N value}
\label{results:single_v_mmo}
\scalebox{0.8}{
\begin{tabular}{*6c}
\toprule
N & SFR-Embedding-Mistral & text-embedding-3-large & bge-large-en-v1.5 & bge-large-en-v1.5-finetuned  & Number of questions \\

1 & 0.7727 & 0.7784 & 0.75577 & \textbf{0.7955}  & 176\\
2 & 0.7143 & 0.6786 & 0.5982 & \textbf{0.8482} & 112\\
3 & 0.3629 & 0.2823 & 0.1774 & \textbf{0.3871} & 124\\
4 & 0.1434 & 0.1434 & 0.0430 & \textbf{0.1720} & 279\\
5 & \textbf{0.0909} & 0.0450 & 0.0227 & 0.0682 & 44\\
6 & \textbf{0.0057} & \textbf{0.0057} & 0.0 & \textbf{0.0057} & 23\\

\midrule
\bottomrule
\end{tabular}}
\end{table}

% From the table 1 we can see the our finetuned model significantly outperformed other baseline models across all metrics. The model outperformed the next best model \textbf{add model name} by over $5 \%$
% in overall hit rate, \textbf{add model name} by over $2 \%$
% in precision@10 and \textbf{add model name} by over $5 \%$
% in recall@10. \newline

From table 1 we can see that our finetuned model significantly outperformed other baseline models across all metrics. The model outperformed the next best model text-embedding-3-large across hit rate, precision@10, and recall@10 quite comfortably,

% \textbf{todo: please explain why the magnitude of precision is lower, recall is higher, and what your accuracy metric means in more detail below} \newline

We see lower precision values in this table due to our choice of $k=10$, since the majority of questions have relevant files at most 4 (more than 91 \%), the number of false positives will be much higher as all 10 files will be included in this evaluation. Similarly, the recall@10 values will be high since $k>n$ will mean more files from our RAG pipeline would be in the universe of relevant files. Our measure of Hit rate@10 is calculated as follows \newline
% \textbf{TODO: revaluate above sentence; make sure it's accurate}

        $   hit_{t}=\left\{
                \begin{array}{ll}
                  1 \; \; \; if \;  files_{n} \subseteq files_{k} \\
                  0 \; \; \; otherwise \\
                \end{array}
              \right. $ \newline 
              
              Where $files_{n}$ is all relevant files for a question and $files_{k}$ are files from the RAG output ($k=10$ in our case)

We also provide the breakdown of hit rate@10 across different values of N; from Table 2 we see the finetuned models outperform other embedding models for almost all values of n except $n=5$ where the SFR-Embedding model performs better. This can be attributed to the large context size and large embedding dimensions of this model which helps in better capturing the embedding of questions requiring a large number of files to solve a particular question.

    % \begin{itemize}
    %     \item \textbf{Analyze the results in detail}
    %     \item \textbf{identify why certain benchmark models fails at more complex tasks}
    %     \item \textbf{Explain why our finetuned embedding model outperformed the benchmarks}
    % \end{itemize}

% \subsubsection{qualitative assessment}

% \begin{itemize}
%     \item \textbf{candidate models and LLMs for benchmarking}
%     \item \textbf{Evaluation dataset: custom dataset}
%     \item \textbf{Evaluation dataset: Can we identify any external evaluation datasets for RAG }
%     \item \textbf{Evaluation metrics: add math notations}
    
% \end{itemize}

{\section{Conclusion}}

In this paper, we outlined why even SOTA embedding models trained for general purpose Retrieval applications fail at task-specific applications. Tabular RAG is a very niche application that is critical for hyper-scaling generative AI applications in the area of data science and finance. SOTA embedding models fail at this application as they are trained on generic datasets, we proposed a novel approach to this problem by finetuning a lightweight BGE-large-en model on our custom dataset. We found that performing new word embedding initialization before finetuning improved the performance of the model. The evaluation results showed that our model outperforms best-in-class embedding models by a significant margin. We posit that for niche task-specific RAG applications finetuning a light embedding model using our approach even on a local laptop would outperform large SOTA embedding models like SFR-Embedding-Mistral and open-AI's ext-embedding-3-large.

\nocite{GPTtech}
\nocite{GPTExperiment}

\bibliographystyle{plainnat}
\bibliography{Report_template}

\end{document}